\documentclass[final,1p,times]{elsarticle}

\usepackage{amssymb}
\usepackage{amsmath}
\usepackage[numbers]{natbib}
\usepackage{float}
\usepackage{multirow}
\usepackage{subfig}
\usepackage[linesnumbered]{algorithm2e}
\usepackage{graphicx}
\usepackage{subcaption}
\usepackage{booktabs}
\usepackage{xcolor}
\usepackage[normalem]{ulem}
\usepackage{xcolor}
\usepackage{hyperref}
\usepackage{wrapfig}

\begin{document}

\begin{frontmatter}



\title{FedCSGA: Evolutionary Client Selection with Joint Statistical and System Heterogeneity in Federated Learning}

\author[label1]{Ghani Haider}
\ead{g.haider@cbnu.ac.kr}

\author[label1]{Majid Kundroo\corref{cor1}}
\ead{kundroomajid@cbnu.ac.kr}

\author[label2]{Leo Zhang}
\ead{leo.zhang@griffith.edu.au}

\author[label3]{Jinchul Choi}
\ead{spiders22v@etri.re.kr}

\author[label1]{Taehong Kim\corref{cor1}}
\ead{taehongkim@cbnu.ac.kr}

\cortext[cor1]{Corresponding authors}

\affiliation[label1]{organization={School of Information and Communication Engineering, Chungbuk National University},
            city={Cheongju},
            country={Republic of Korea}}

\affiliation[label2]{organization={School of Information and Communication Technology, Griffith University},
            city={Gold Coast},
            country={Australia}}

\affiliation[label3]{organization={Electronics and Telecommunications Research Institute},
            city={Daejeon},
            country={Republic of Korea}}

\begin{abstract}
Federated Learning (FL) enables collaborative training of machine learning models across distributed clients while preserving data privacy. In practical deployments, FL relies on selecting a client subset in each communication round to reduce communication cost, mitigate stragglers, and improve training efficiency under heterogeneous data and system conditions. In heterogeneous edge computing environments, both statistical heterogeneity and system heterogeneity significantly affect convergence efficiency and model performance. However, existing client selection methods often address these factors independently, overlooking their coupled and dynamic impact across communication rounds.
To address this limitation, we propose \textbf{FedCSGA}, a \textbf{C}lient \textbf{S}election method for \textbf{Fed}erated learning based on \textbf{G}enetic \textbf{A}lgorithm. FedCSGA encodes client subsets as chromosomes and evaluates these subsets through a non-linear fitness function that jointly integrates resource-aware metrics and history-aware participation weighting. This design captures subset-level interactions, mitigates persistent selection bias, and balances statistical utility with system efficiency.
Extensive experiments on FMNIST, CIFAR-10, CIFAR-100, and Tiny-ImageNet demonstrate that FedCSGA consistently outperforms state-of-the-art baselines across diverse heterogeneity settings, achieving accuracy improvements of up to $7.0\%$ and faster convergence with fewer communication rounds. These results validate the effectiveness of jointly modeling statistical and system factors to achieve robust and efficient learning in heterogeneous federated environments.
\end{abstract}

\begin{keyword}
Federated learning \sep Genetic algorithm \sep Client selection \sep Statistical heterogeneity \sep System heterogeneity \sep Edge computing

\end{keyword}

\end{frontmatter}

\def\thefootnote{}\footnotetext{\textcopyright~2026. This manuscript version is made available under the CC-BY-NC-ND 4.0 license (\url{http://creativecommons.org/licenses/by-nc-nd/4.0/}). This is the accepted manuscript version of an article published in \textit{Journal of Systems Architecture}. The final published version is available at \url{https://doi.org/10.1016/j.sysarc.2026.103813}.}
\def\thefootnote{\arabic{footnote}}

\section{Introduction}\label{sec:introduction}
Federated Learning (FL) has emerged as a transformative paradigm for privacy-preserving distributed machine learning \cite{uddin2025systematic}. Its deployment increasingly spans heterogeneous edge computing environments, where mobile, IoT, and edge–cloud continuum systems jointly participate in model training. Such environments impose constraints shaped by real-time processing demands, limited resources at edge clients, and dynamic network conditions \cite{houidi2025evaluating}. Due to resource limitations and privacy requirements, FL typically selects only a subset of available clients in each communication round, allowing these clients to perform local model updates without sharing raw data \cite{houidi2024federated}. However, partial client participation introduces significant challenges arising from both statistical and system heterogeneity \cite{benarba2025bias,SELO2026108177}.

Statistical heterogeneity arises when client datasets vary in size and originate from diverse distributions, resulting in non-Independent and Identically Distributed (non-IID) data \cite{solans2024non}. This phenomenon can degrade global model performance, increase gradient variance, and slow convergence \cite{YU2024103280}. System heterogeneity refers to variations in computational resources, communication bandwidth, and processing capabilities across clients \cite{yang2025coordinating}. These disparities often lead to straggler effects, where slower or resource-constrained clients delay global synchronization and reduce training efficiency \cite{wong2025empirical}. While random client selection remains the predominant strategy in practice, it fails to explicitly account for these heterogeneities, frequently resulting in inefficient resource utilization and unstable convergence behavior \cite{KANG2025103575}.

To address these challenges, recent studies have explored client selection strategies that explicitly consider either statistical utility or system characteristics \cite{mayhoub2024review,ZHAO2025103576}. Some approaches prioritize clients based on data-related metrics, such as loss, gradient norms, or data quality \cite{cho2020client,goetz2019active}, while others emphasize system-level factors, including computational capacity \cite{fedprox_li,ZHANG2025103526} or communication constraints \cite{singhal2024greedy}. However, most existing methods address these heterogeneity factors in isolation and fail to capture the client subset-level statistical diversity and system compatibility \cite{ahmed2025adaptive}, particularly under repeated participation across rounds.

In this work, we propose FedCSGA, a client selection framework for federated learning that leverages the Genetic Algorithm (GA) to select an optimal subset of participating clients in each training round. In FedCSGA, each candidate solution encodes a fixed-size subset of participating clients as a chromosome, and its quality is evaluated by a fitness value that jointly considers statistical properties (e.g., data quality) and system resources (e.g., computational power). Importantly, the fitness function incorporates a history-aware participation weighting mechanism that introduces subset-level and cross-round dependencies among clients. This creates an inherently non-additive selection problem unsuitable for simple greedy ranking. Through iterative evolutionary operations – including selection, crossover, and mutation – FedCSGA explores the combinatorial search space to identify optimal client subsets that balance model contribution and system efficiency. This design overcomes the limitations of existing selection methods and provides a robust subset-level strategy for adaptive client participation in heterogeneous FL settings. The main contributions of this work are summarized as follows:

\begin{itemize}
\item We propose FedCSGA, a GA-based client selection framework that formulates client selection as a subset-level combinatorial problem, explicitly addressing both statistical and system heterogeneity in federated learning.
\item We design a fitness function that integrates resource-aware metrics with history-aware participation weighting, capturing non-linear and cross-round dependencies and enabling balanced and efficient client selection.
\item Extensive experiments on FMNIST, CIFAR-10, CIFAR-100, and Tiny-Imagenet demonstrate that FedCSGA improves test accuracy by an average of $1.0-7.0\%$ over strong baselines across both IID and non-IID settings, while maintaining stable convergence and efficiency.
\end{itemize}

To ensure full reproducibility of our framework, the source code for FedCSGA is publicly available at \url{https://github.com/nclabteam/FedCSGA}.

\section{Background}\label{sec:background}
This section reviews the fundamental concepts of FL, the role of client selection, and the principles of GA as heuristic search mechanisms for combinatorial problems. These components form the methodological foundation of our proposed approach.

\subsection{Federated Learning}
FL is a distributed machine learning paradigm in which a global model is collaboratively trained across a set of decentralized clients, each holding private local data that cannot be shared directly due to privacy, security, or regulatory constraints \cite{mcmahan2017communication,uddin2025systematic}. In a typical FL workflow, a central server initializes a global model and orchestrates training over multiple communication rounds. In each round, a client subset downloads the current global model, performs local training on private data, and uploads model updates to the server, where they are aggregated to update the global model \cite{houidi2024federated}.

A defining characteristic of FL is partial participation, where only a fraction of the available clients participate in any given communication round. Partial participation is necessary due to device availability, system limitations, and straggler effects in large-scale deployments \cite{houidi2025evaluating,wong2025empirical}. While it improves scalability, partial participation introduces challenges related to convergence efficiency and model generalization, particularly in heterogeneous environments \cite{benarba2025bias,SELO2026108177}.

FL systems are commonly affected by two forms of heterogeneity. Statistical heterogeneity arises when client datasets differ in size, class distribution, or feature characteristics, leading to non-IID data across clients \cite{solans2024non}. System heterogeneity refers to variations in computational capability, memory, energy constraints, and communication latency \cite{yang2025coordinating}. These heterogeneities jointly influence training stability, convergence behavior, and system efficiency, making coordination among clients a central challenge in FL \cite{YU2024103280}.

\subsection{Client Selection in Federated Learning}
Client selection is a core mechanism in FL that determines which clients participate in each communication round. Given a pool of available clients, the server must select a subset that balances learning performance, training efficiency, and system constraints. Early FL frameworks, such as FedAvg \cite{mcmahan2017communication,CAO2022102413}, adopt uniform random client selection. While simple and unbiased, random selection ignores client-level differences in data utility and system resources, often resulting in slow convergence and inefficient resource utilization in heterogeneous settings \cite{CHOI2025103569,KANG2025103575}.

The primary objective of client selection is to mitigate the adverse effects of statistical and system heterogeneity while maintaining scalability. From a statistical perspective, selecting clients with informative, diverse, or representative data can improve gradient quality and accelerate convergence \cite{goetz2019active,cho2020client}. From a system perspective, avoiding resource-constrained or unreliable clients can reduce straggler delays and communication overhead \cite{nishio2019client,ali2025fedheur}. These objectives, however, are often conflicting, as clients with valuable data may exhibit limited system capabilities, whereas resource-rich clients may contribute less informative updates \cite{mayhoub2024review,ZHAO2025103576}.

Consequently, client selection can be formulated as an optimization problem where the goal is to identify a client subset that provides a favorable trade-off between learning utility and system efficiency. The search space grows combinatorially with the number of clients, rendering exhaustive search infeasible in practical FL deployments \cite{singhal2024greedy}. This motivates the use of heuristic search methods that can efficiently explore the space of client subsets under multiple competing criteria.

\subsection{Genetic Algorithms}
Genetic Algorithms are population-based, stochastic evolutionary optimization methods inspired by the principles of natural evolution, including selection, crossover, and mutation \cite{holland1992genetic}. A GA maintains a population of candidate solutions, referred to as chromosomes, each encoding a potential solution to the optimization problem. The quality of each chromosome is evaluated using a fitness function, which guides the selection of candidates for reproduction across successive generations. Through iterative application of genetic operators, GAs balance exploration of the search space with exploitation of promising solutions, making them suitable for efficiently approximating an optimal solution within complex and combinatorial search spaces.

These properties make GAs a natural fit for FL client selection, where the utility of a selected subset depends not only on individual client characteristics but also on interactions among clients in terms of statistical diversity and system compatibility \cite{ahmed2025adaptive}. By encoding client subsets as chromosomes and evaluating them through a fitness function that reflects multiple selection criteria, GAs provide a flexible framework for approximating effective client selection strategies in heterogeneous FL environments.

\section{Related Work}\label{sec:related_work}
In the traditional FL algorithms, the client subset for training is selected uniformly at random and independently at each iteration~\cite{CAO2022102413}. While some algorithms rely on regularization terms~\cite{karimireddy2020scaffold,fedprox_li} and aggregation methods~\cite{jhunjhunwala2022fedvarp} to overcome the heterogeneity challenges of clients, they often fail to account for differences in data distributions and system capabilities across clients. This often results in slow convergence and inefficient resource utilization in heterogeneous environments~\cite{CHOI2025103569}.

Complementary to these approaches, client selection algorithms strategically choose clients for training in each round to mitigate statistical or system heterogeneity and enhance global model accuracy. To mitigate statistical heterogeneity, several approaches prioritize clients based on data-related utility. Active Federated Learning (AFL)~\cite{goetz2019active} assigns higher importance to clients with larger training losses, under the assumption that such clients contribute more informative updates. The Power-of-Choice (PowD)~\cite{cho2020client} strategy refines this by selecting the highest-loss clients from randomly sampled subsets, improving convergence behavior under non-IID data distributions. Contribution-aware methods, such as Greedy Shapley (GreedyFed)~\cite{singhal2024greedy}, model client participation as a cooperative game and rank clients based on their marginal contribution to validation performance. Recent studies have also introduced diversity-aware~\cite{YU2024103280} or submodular optimization frameworks~\cite{balakrishnan2022diverse} to select representative client subsets that maximize information gain. While effective in improving statistical efficiency, these methods impose strict theoretical assumptions (e.g., the availability of an unbiased server-side validation dataset) and often incur substantial computational overhead due to repeated evaluation of client contributions.

In contrast, system and resource-aware client selection strategies focus on improving training efficiency by accounting for computation latency, energy consumption, or communication cost. Specifically, methods like FedCS~\cite{nishio2019client} and FedHeur~\cite{ali2025fedheur} are designed for mobile edge computing environments to select clients based on training time or task offloading considerations. However, these methods trade statistical representation for system speed, often leading to biased models under non-IID conditions. More recently, Reinforcement Learning (RL)-based methods~\cite{ahmed2025adaptive} have been proposed to adapt to dynamic heterogeneous environments. While these approaches learn selection policies that balance accuracy and latency over time, they require extensive warm-up training rounds, assume stationary network environments, and struggle to scale without severe state-space explosion.

As shown in Table~\ref{tab:client_selection_related_work}, existing literature reveals a persistent trade-off between statistical representation, system constraints, and computational scalability. Methods addressing statistical and system heterogeneities either rely on idealized assumptions (e.g., proxy datasets, perfectly synchronous reporting) or evaluate clients as independent entities rather than evaluating statistical diversity and system compatibility at the client subset level. FedCSGA differs from prior work by formulating client selection as a combinatorial optimization problem over entire client subsets and employing GA as a metaheuristic search strategy. Unlike greedy baselines or single-criterion heuristics, the proposed approach evaluates candidate subsets using a non-linear fitness function that jointly reflects statistical contribution and system resource constraints. This enables FedCSGA to approximate balanced client selections that improve convergence behavior and training efficiency in heterogeneous FL settings.

\begin{table}[!t]
\centering
\caption{Comparison of Client Selection Methods in Federated Learning}
\label{tab:client_selection_related_work}
\resizebox{\textwidth}{!}{
\begin{tabular}{p{2.4cm} p{2.2cm} p{2.2cm} p{2.8cm} p{2.6cm} p{4.0cm}}
\hline
\textbf{Work} 
& \textbf{Strategy} 
& \textbf{Heterogeneity}
& \textbf{Key Assumptions} 
& \textbf{Overhead} 
& \textbf{Limitations} \\
\hline

FedAvg \cite{mcmahan2017communication} 
& Random sampling 
& None 
& Uniform client availability and reliability
& Very low (random sampling over clients)
& Ignores statistical and system heterogeneity; slow convergence under non-IID. \\

AFL \cite{goetz2019active} 
& Loss-based prioritization 
& Statistical  
& Local training loss reflects data informativeness
& Low (requires per-round loss evaluation)
& Neglects system constraints; sensitive to unstable loss signals. \\

PowD \cite{cho2020client} 
& Greedy (subset sampling and top-k loss) 
& Statistical 
& High-loss clients contribute more to global improvement
& Low (selection within sampled subset)
& Ignores system heterogeneity; may over-select noisy clients. \\

GreedyFed \cite{singhal2024greedy} 
& Marginal contribution (Shapley-based) 
& Statistical 
& Availability of unbiased server-side validation data
& High (repeated validation-based contribution evaluation)
& Requires proxy validation data; greedy selection with limited exploration. \\

FedCS \cite{nishio2019client}
& Greedy scheduling (time-constrained)
& System
& Accurate estimation of computation and communication latency
& Low (greedy selection based on resource constraints)
& Requires accurate resource info; ignores statistical heterogeneity. \\

RL-based \cite{ahmed2025adaptive} 
& Policy learning
& Statistical and System
& Stationary environment and well-designed state/reward representation
& High (policy training and state-space modeling)
& Significant training overhead; scalability limited by state-space complexity. \\

\hline
\end{tabular}
}
\end{table}

\section{Proposed method}\label{sec:method}
We propose FedCSGA, a resource-aware client selection algorithm based on GA, designed to address both statistical and system heterogeneity in FL. The key mathematical symbols used throughout the formal description are summarized in Table~\ref{tab:notations}.

\begin{table}[!t]
    \small
    \setlength{\tabcolsep}{20pt} 
    \renewcommand{\arraystretch}{0.9} 
    \centering
    \caption{Frequently used notations in FedCSGA}
    \label{tab:notations}
    \begin{tabular}{ll}
        \toprule
        \textbf{Symbol} & \textbf{Description} \\
        \midrule
        $w$ & global model parameter in $\mathbb{R}^d$ \\
        $F(w)$ & global objective function\\
        $F_{i}(w)$ & local objective function of the $i$-th client \\
        $D_i$ & local dataset of the $i$-th client \\
        $\delta_{j}$ & a data sample from the local dataset $D_i$ \\
        $L(w, \delta_{j})$ & loss function with parameter \(w\) and data sample \( \delta_{j} \) \\
        $\hat{L}(w, \delta_{j})$ & normalized loss over all data samples of a client \\
        $T$ & total number of training rounds \\
        $C$ & set of all available clients \\
        $K$ & total number of clients\\
        $w^t_i$ & local model update of the $i$-th client in round $t$ \\
        $w^{t+1}$ & updated global model parameter at round $t+1$ \\
        \(\alpha\) & dirichlet concentration parameter \\
        \(\eta\) & learning rate of the local model \\
        $S_t$ & subset of $m$ selected clients in round $t$ \\
        $\mathcal{S}$ & search space for all possible client subsets $S_t$ \\
        $S_t^*$ & selected client subset in round $t$ \\
        $W(c_i)$ & history-aware selection weighting factor\\
        $\text{Res}(c_i)$ & total resources of client $c_i$ \\
        $\text{Res}_{\text{stat}}(c_i)$ & statistical resources of client $c_i$ \\
        $\text{Res}_{\text{sys}}(c_i)$ & system resources of client $c_i$ \\
        $f_{c_i}$ & cumulative selection count of client $c_i$ \\
        $\gamma$ & decay rate for history-aware weighting \\
        $\tau_{i}$ & compute time of client $c_i$ for a local model update \\
        $\lambda_r$ & weighting factor for statistical resources \\
        $\lambda_s$ & weighting factor for system resources \\
        $\text{Fitness}(S_t)$ & fitness function of a subset $S_t$\\
        $P$ & population size of the GA \\
        $G$ & number of generation rounds of the GA \\
        $\rho_p$ & parent chromosomes selection fraction \\
        $\rho_c$ & child chromosomes selection fraction \\
        \bottomrule
    \end{tabular}
\end{table}

\subsection{Problem formulation}
FL aims to optimize the global model parameters \( w \in \mathbb{R}^d \) by minimizing the global objective function \( F(w) \), which represents the aggregated performance across all \( K \) clients. The global objective function is defined as:
\begin{equation}
F(w) = \frac{1}{K} \sum_{i=1}^{K} F_{i}(w),
\end{equation}
where \( F_{i}(w) \) denotes the local objective function of the \( i \)-th client, evaluated over its local dataset \( D_i \),
\begin{align}
F_{i}(w) = \frac{1}{|D_{i}|} \sum_{\delta_{j} \in D_{i}} L(w, \delta_{j}).
\end{align}

Here, \( L (w, \delta_{j}) \) represents the loss function with \(w \in \mathbb{R}^d\), and \( \delta_{j} \) denotes a data sample drawn from the local data distribution \( D_i \). FL often involves partial client participation in each training round due to statistical heterogeneity and system resource constraints. Let the set of all available clients be \( C = \{c_1, c_2, \ldots, c_K\} \). In each round \( t \), a subset \( S_t \) of \( m \) clients (\( |S_t| = m \leq |C| \)) is selected from the search space \( \mathcal{S} \), defined as:
\begin{align}
\mathcal{S} = \{ S_t \mid S_t \subset C , |S_t| = m \}.
\end{align}
The objective is to minimize \( F(w) \) by selecting the highest quality subset \( S_t^* \) in each round by maximizing a fitness function that combines resource utility and history-aware participation weighting:
\begin{align}
S_t^* = \arg\max_{S_t \in \mathcal{S}} \Bigg(\sum_{c_i \in S_t} \text{Res}(c_i)\Bigg) \cdot \frac{1}{|S_t|} \sum_{c_i \in S_t} W(c_i).
\end{align}
This formulation is non-linear and non-additive, capturing the complex, subset-level interactions between clients.
Here, \( W(c_i) \) is a history-aware selection weighting factor defined as
\begin{align}
W(c_i) = e^{-\gamma f_{c_i}},
\end{align}
and \( \text{Res}(c_i) \) is the resource metric of client \( c_i \), comprising both statistical and system contributions
\begin{align}
\text{Res}(c_i) = \lambda_{r} \text{Res}_{\text{stat}}(c_i) + \lambda_{s} \text{Res}_{\text{sys}}(c_i).
\end{align}

Here, \( f_{c_i} \) denotes the cumulative number of rounds in which client \( c_i \) has been selected, and \( \gamma \) is a decay rate that controls how strongly historical participation reduces the future selection weight of client \( c_i \). \( \text{Res}_{\text{stat}}(c_i) \) quantifies the statistical resource contribution of client \( c_i \), while \( \text{Res}_{\text{sys}}(c_i) \) represents its system-level resources. The weighting factors \( \lambda_{r} \) and \( \lambda_{s} \) control the influence of statistical and system resources, respectively. To maximize the history-aware resource utility in each communication round, we employ GA to select the best-performing subset \( S_t^* \) from the search space \( \mathcal{S} \). The GA initializes and evolves a population of candidate subsets \( S_t^j \) (\( j = 1, 2, \dots, |P| \)) over \( G \) generations, jointly accounting for client resource contributions and historical selection pressure. The algorithm ultimately converges to the subset \( S_t^* \) that achieves the highest history-aware resource utility, which is returned as the final client subset for the round.

\subsection{Federated client selection using genetic algorithm}
Our FedCSGA algorithm leverages client resource information for client selection, ensuring the participation of high-quality clients in each FL round. As shown in Figure~\ref{fig:fedcsga_framework}, each client transmits its resource metrics to the server, which then uses GA for client selection, involving initialization, reproduction, and population updates. During reproduction, client subsets generated through parent selection, crossover, and mutation are evaluated over multiple generations to identify the highest quality client subset \(S_t^*\). This process selects clients with high resource availability, improves model accuracy, and accelerates global model convergence.

Algorithm~\ref{alg:fedcsga} outlines the detailed steps of the procedure. At the beginning of each training round \(t\), as described in line \(5\) of Algorithm~\ref{alg:fedcsga}, each client transmits its resource metrics (\(\text{Res}\)) to the central server. The aim is to select \(m\) clients for training using the client selection function specified in line \(6\). Lines \(7 \text{--} 13\) follow the standard procedure of FL, involving local training by selected clients and global model aggregation at the server. For client selection, the resource information collected from clients at the server is utilized in the client selection function (line \(6\)) to initialize the GA generations, which perform the selection of \(m\) clients for the current round. The process consists of the following steps:

\begin{figure*}[!t]
    \centering
    \includegraphics[width=\textwidth]{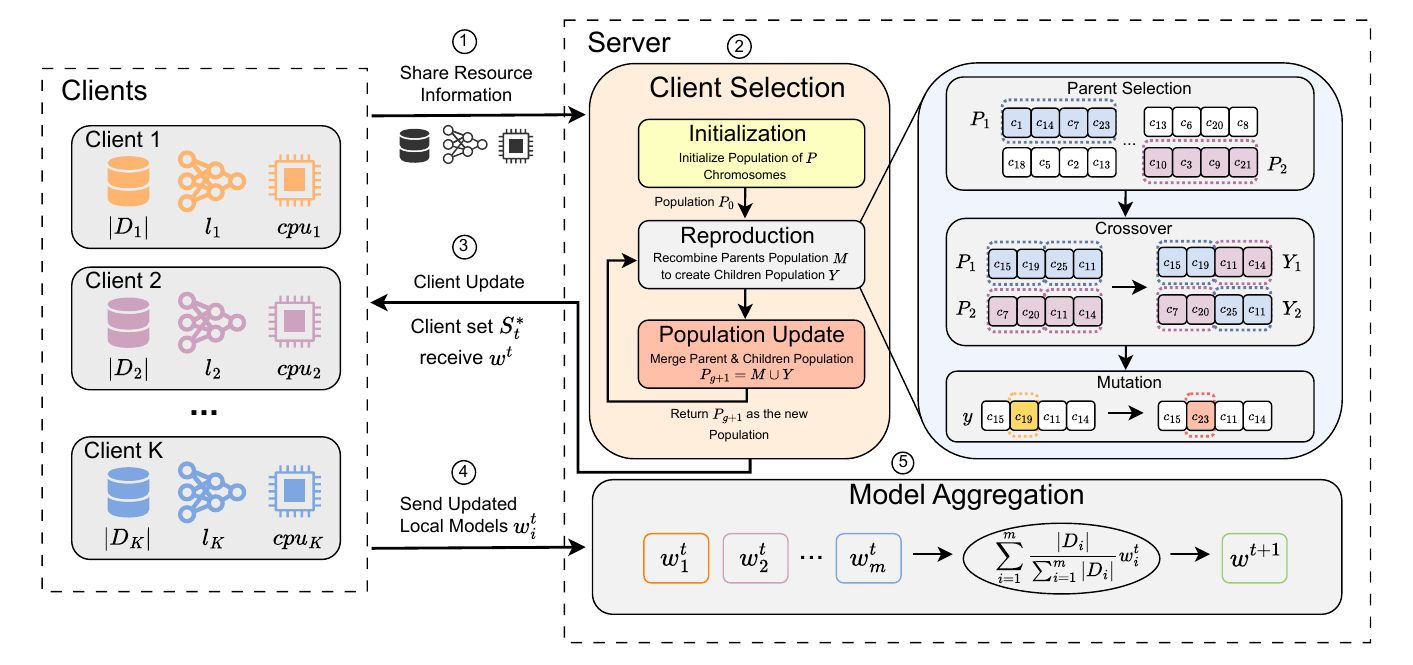}
    \caption{Overview of FedCSGA algorithm for each training round.}
    \label{fig:fedcsga_framework}
\end{figure*}

\SetKwComment{Comment}{/* }{ */}
\RestyleAlgo{ruled}
\begin{algorithm}[!t]
\small
    \caption{FedCSGA}\label{alg:fedcsga}
    \textbf{Input:}
        set of all available clients \( C \),
        number of communication rounds \(T\),
        number of participants \(m\),
        number of total GA generations \(G\), 
        number of total chromosomes \(P\), 
        parent chromosomes fraction \( \rho_p \),
        child chromosomes fraction \( \rho_c \),
        client resources \( \text{Res} \),
        statistical resource weight \( \lambda_{r} \),
        system resources weight \( \lambda_{s} \)

    \textbf{Initialize:} global model \(w^{0}\)\;
    \textbf{Server Executes:} \\
    \For{\textnormal{each round} \(t = 1, 2, \dots , T\)}{
        \( \text{Res} \gets \) resource metric \(\text{Res}(c)\) from all clients \(c\) in \( C \)\;
        \(S_{t}^{\ast} \gets \textbf{ClientSelection}(C, m, \text{Res})\)\;
        \For{\(i \in S_{t}^{\ast} \textnormal{\textbf{ in parallel}}\)}{
        send \(w^{t}\) to client \(i\) and initiate local training\;
        }
        \( w_{1}^{t}, \cdots, w_{m}^{t} \gets \) collect \(w_{i}^{t}\) from clients\;
        \( w^{t+1} \gets \sum_{i=1}^{m} \frac{|D_{i}|}{\sum_{i=1}^{m}|D_{i}|} w_{i}^{t} \)\;
    }
    \textbf{return} \( w^{T} \)\;
    
    \BlankLine 
    \BlankLine 
    \textbf{ClientSelection}\( (C, m, \text{Res}) \)\textbf{:} \\
    \( P_{0} = \{S_t^1, S_t^2, \dots, S_t^P\} \gets \) initialize each chromosome \( S_t^j \) with \( \{c_i \mid c_i \in C, \, i = 1, 2, \dots, m\} \)\;
    \For{\(g = 1, 2, \dots , G \)}{
        \tcp{Parent Selection}
        \( M \gets \) choose \( \rho_p |P_{g}| \) chromosomes \(S_t^j\) from \(P_{g}\) in descending order of \(\textbf{Fitness}(S_t^j)\) \;
        \tcp{Crossover}
        \For{\(q = 1, 2, \dots , \frac{\rho_c |P_{g}|}{2} \)}{
            \(P_{1}, P_{2} \gets\) select the two highest-fitness chromosomes from a tournament of size \( \rho_c |P_g| \) sampled from \( M \)\;
            $L \gets$ A random index $L$ is sampled uniformly from $[1, |S_t^j|]$\;
            \(Y_{1}, Y_{2} \gets\) $L$ prefix-based crossover preserving client uniqueness and subset cardinality \(m\)\;
            \( Y \gets \{ Y_{1}, Y_{2} \} \) \;
        }
        \tcp{Mutation}
        \For{\textnormal{each} \( y \in Y \)}{
            \For{\textnormal{each} \( c_{i} \in y \)}{
                replace \( c_{i} \) with probability \( \frac{1}{|y|} \) \;
            }
        }
        \tcp{Population Update}
        \( P_{g+1} \gets M \cup Y \) \;
    }
    \textbf{return} \( S_t^* \gets \operatorname*{arg\,max}_{S_t^j \in P_{G}} \textbf{ Fitness}(S_t^j) \)\;
    
    \BlankLine 
    \textbf{Fitness}\( (S_t^j) \)\textbf{:} \\
    \textbf{return} \( \left( \sum_{c_i \in S_t^j} \text{Res}(c_i) \right) \cdot \frac{1}{|S_t^j|} \sum_{c_i \in S_t^j} e^{-\gamma f_{c_i}} \);
\end{algorithm}

\subsubsection{Initialization}
The GA initializes a population \( P_{0} \) of \( P \) chromosomes with \(\{c_i \mid c_i \in C, \, i = 1, 2, \dots, m\}\) in line \( 15 \), where each chromosome \(S_t^j\) (\( j = 1, 2, \dots, |P| \)) represents a subset of $m$ distinct clients selected from the available client pool \(C\). Chromosomes are initialized by uniformly sampling $m$ clients without replacement, ensuring that no duplicate clients appear within a subset by construction.

Each client $c_i$ is associated with a scalar resource score $\text{Res}(c_i)$ that captures its expected contribution to the current training round. $\text{Res}(c_i)$ is computed as a weighted combination of client-specific statistical \( \text{Res}_{\text{stat}} \) and system \( \text{Res}_{\text{sys}} \) metrics available at the server. By integrating both statistical and system-aware components, the fitness function guides the GA toward client subsets that enhance both learning effectiveness and system efficiency. The fitness of a candidate subset $S_t^j$ is defined as
\begin{equation}
\text{Fitness}(S_t^j) 
= \left( \sum_{c_i \in S_t^j} \text{Res}(c_i) \right)
\cdot
\frac{1}{|S_t^j|}
\sum_{c_i \in S_t^j} e^{-\gamma f_{c_i}},
\label{eq:fitness}
\end{equation}
where $f_{c_i}$ denotes the cumulative selection count of client $c_i$ up to round $t$, and $\gamma > 0$ is the decay rate for history-aware weighting.

This formulation introduces an explicit coupling between clients within a subset. Although statistical and system resources are linearly combined at the client level, their joint effect on selection emerges at the subset level through the non-linear history-aware weighting, preventing dominance by any single resource dimension. As a result, the fitness of a subset depends not only on the individual quality of its members but also on their collective selection history, discouraging repeatedly selecting the same group of high-resource clients and promoting long-term participation balance.

\subsubsection{Reproduction}
The GA evolves the population over multiple generations to improve the fitness of chromosomes. At each generation, the population is updated using elitist survival and stochastic reproduction. A fixed proportion of the top-performing chromosomes is preserved as parents to the next generation to maintain an optimal solution. The remaining offspring are generated via tournament-based parent selection, which balances exploitation of high-fitness subsets with population diversity. Lines \( 16 \text{ - } 28\) explain the reproduction process where three main operations are performed: \textit{parent selection}, \textit{crossover}, and \textit{mutation}.
\begin{itemize}
    \item Parent Selection: In line \( 17 \), \( \rho_p|P_g| \) chromosomes with the highest fitness values from the current population \(P_g\) are selected as parents, where \( \rho_p \) determines the fraction of chromosomes retained from the previous generation as parents. These parents serve as the basis for creating new child chromosomes with potentially better fitness values.
    \item Crossover: Lines \( 18 \text{ - } 23\) explain the crossover operation. Two parent chromosomes \( P_1 \) and \( P_2 \) are selected via tournament selection with the highest fitness value among \( \rho_c|P_g| \) number of chromosomes in \(M\), where \( \rho_c \) specifies the fraction of new chromosomes generated through crossover. A random index \( L \) is sampled uniformly from \( [1, |S_t^j|] \). Since chromosomes encode fixed-cardinality client subsets rather than independent genes, naive crossover operators may yield infeasible or redundant selections. Moreover, an optimal client subset in FL often exhibits common structural characteristics such as complementary data distributions or favorable resource availability, motivating a recombination-based operator. We therefore adopt a crossover operation that ensures client uniqueness without requiring post-hoc repair. Specifically, a prefix of length \( L \) is inherited from the first parent. The remaining positions are filled using the second parent, excluding duplicates. This process is symmetrically applied to generate two child chromosomes, \( Y_1 \) and \( Y_2 \), ensuring that each offspring represents a valid subset of \( m \) distinct clients.
    \item Mutation: For each child chromosome \( y \), every client in \( y \) is subject to a mutation process with a low probability of \( 1/|y| \) in lines \( 24 \text{ - } 28 \). If a mutation occurs, the corresponding client in \( y \) is replaced with another random client from \( C \) that does not already appear in \( y \). This simple mutation operator preserves subset cardinality and feasibility while introducing controlled diversity into the population. While crossover exploits existing high-fitness client combinations, mutation enables exploration of previously unselected clients and mitigates premature convergence, which is particularly important under non-IID data distributions commonly encountered in FL.
\end{itemize}
The combined use of crossover and mutation allows the GA to adapt client selection dynamically across rounds, reflecting the evolving contribution of clients as local models and data distributions change. We deliberately employ simple, low-overhead GA operators rather than more complex adaptive or problem-specific variants. This design choice reflects the practical constraints of FL systems, where client selection must be performed efficiently at the server without introducing significant computational overhead or instability across rounds. 

\subsubsection{Population Update}
After crossover and mutation, the population is updated in line \( 29 \) to \( P_{g+1} \) in order to include both parent and child chromosomes for the next GA generation. This ensures that high-fitness chromosomes are retained while introducing diversity into the population. The reproduction process is repeated for \( G \) generations or until no significant improvement in the fitness of chromosomes is observed, after which the chromosome with the highest fitness value is selected as the final solution in line \( 31 \) and returned as \( S_{t}^{*} \).

\subsubsection{Client Selection and Training}
The best chromosome obtained from the GA represents the subset \( S_t^* \) of clients for the current round \(t\). The server selects these clients to participate in local training. In lines \(7 \text{ - } 10\), each selected client receives the current global model \( w^t \), performs local training using its dataset \( D_{i} \), and sends back the updated model \( w^t_i \). The central server aggregates these updates to produce the new global model:
\begin{align}
w^{t+1} \gets \sum_{i=1}^{m} \frac{|D_{i}|}{\sum_{i=1}^{m}|D_{i}|} w_{i}^{t}
\end{align}

To enable history-aware selection, the server maintains and updates the selection frequency $f_{c_i}$ for each client across communication rounds. After each round, the frequency of selected clients is incremented, and the updated values are used in subsequent rounds when evaluating the history-aware decay term in Eq.~\eqref{eq:fitness}. The entire process is repeated for a predefined number of FL rounds or until the global objective \( F(w) \) satisfies a specified convergence criterion. By using GA, the client selection process becomes resource-aware, leading to enhanced global model performance.

\subsubsection{Design of Resource Metrics} 
Prior works \cite{yang2025heaps, balakrishnan2022diverse} have explored strategies where clients with higher statistical utility are prioritized based on importance sampling, gradient norm, or weight differences between client local models and the global model. However, directly computing gradient norms or changes in the local model of clients in each round introduces significant computational overhead, making it impractical in resource-constrained environments. To address this issue, we leverage insights from \cite{cho2020client}, which demonstrate that choosing clients with higher local loss accelerates convergence. Consequently, we approximate statistical heterogeneity through a pragmatic metric that incorporates both the client’s dataset size and its most recent training loss, thereby quantifying its statistical contribution with minimal computational overhead. Specifically, we define the statistical resource metric of client \( c_i \) as:
\begin{align}
\text{Res}_{\text{stat}}(c_i) = {|D_i|} \cdot \sum_{\delta_j \in D_i} \hat{L}(w, \delta_j),
\end{align}
where \( \hat{L}(w, \delta_j) \) represents the normalized loss over all local training samples of client \( c_i \) at its last participation round. This formulation enables the exploitation of clients with larger datasets and higher recent losses, thereby facilitating faster convergence. For clients that have not yet participated, \( \hat{L}(w, \delta_j) \) is initialized to a large constant, encouraging early exploration of previously unseen clients. Once selected, this value is updated based on the observed local training loss and, together with the history-aware selection mechanism, promotes training stability and fairness.

For system heterogeneity, we follow a similar setup of \cite{reisizadeh2022straggler, fei2025addressing}, recognizing that federated clients exhibit diverse computational capabilities, which introduce variations in local model update times. To formally characterize this heterogeneity, we define \( \tau_i \) as the expected time required for client \( c_i \) to complete a single local model update. This duration is predominantly influenced by the computation time associated with processing a fixed batch-size stochastic gradient of the local empirical loss function \( L_i(w) \) \cite{reisizadeh2022straggler}. Clients with larger \( \tau_i \) values correspond to lower computational efficiency, often functioning as stragglers that hinder synchronization and overall training progression. To systematically capture system heterogeneity, we define the system resource metric as follows:
\begin{align}
\text{Res}_{\text{sys}}(c_i) = \frac{1}{\tau_{i}},
\end{align}
This formulation ensures that clients with lower computational latency contribute higher system resource values, facilitating the selection of more efficient clients for training.

Both the statistical and system resource metrics proposed in this work are intentionally designed to be modular and extensible, facilitating their adaptation to more nuanced FL environments. For statistical heterogeneity, although we currently leverage dataset size and recent training loss due to their low overhead, the formulation permits integration of richer data characteristics such as sample diversity \cite{li2021sample}, label distribution imbalance \cite{duan2019astraea}, or noise levels\cite{chen2020dealing}. On the system side, the current design captures computational latency via local update time, but can be augmented to account for communication-related overheads, including bandwidth variability \cite{xu2020client}, transmission latency \cite{al2023fedratrees}, or energy consumption \cite{yang2020energy}. For instance, an extended formulation could incorporate both computation and communication time through a unified delay-aware system resource metric. Overall, this design maintains computational efficiency while providing a principled foundation for future integration of statistically and systemically richer client attributes, thus enhancing selection decisions under real-world FL heterogeneity. 

\section{Evaluation}\label{sec:exp_setup}
This section outlines the experimental setup, including baselines, datasets, model architectures, and training configurations. We then evaluate FedCSGA through extensive experiments encompassing performance comparisons, partial client participation analysis, GA parameter study, ablation experiments, and communication and computation complexity assessment.

\subsection{Experimental setup}\label{subsec:exp_setup}

\subsubsection{Baselines} To demonstrate the effectiveness of FedCSGA, we compared it against a diverse set of baseline algorithms. These baselines include both traditional FL and client selection algorithms.
\begin{itemize}
    \item \textbf{\textit{FedAvg}} \cite{mcmahan2017communication} The standard FL aggregation algorithm that performs a weighted averaging of local model updates from randomly selected clients.
    \item \textbf{\textit{FedProx}} \cite{fedprox_li} An FL algorithm that allows each participating device to perform a variable amount of work by stabilizing local training with a proximal term \(\mu\). The parameter \(\mu\) in FedProx was tuned from \(\{0.1, 0.5, 1.0\}\).
    \item \textbf{\textit{PowD}} \cite{cho2020client} A client selection algorithm that prioritizes clients with the highest losses from a randomly sampled subset to train in FL. The parameter \(d\) in PowD where \((m \leq d \leq K\)) was tuned from \(\{15, 20, 25, 35\}\).
    \item \textbf{\textit{AFL}} \cite{goetz2019active} An FL algorithm that selects clients with a probability conditioned on the current model and the data on the client. The parameters \(\alpha_1\), \(\alpha_2\), and \(\alpha_3\) in AFL were set to their default values of \(0.75\), \(0.01\), and \(0.1\), respectively.
    \item \textbf{\textit{GreedyFed}} \cite{singhal2024greedy} A greedy client selection algorithm that uses Shapley value from cooperative game framework to measure client contribution. The parameters \(T\) and \(\epsilon\) in GreedyFed were set to their default values of \(50 \times m\) and \(0.0001\), respectively.
    \item \textbf{\textit{GreedyFL}} A greedy client selection strategy that utilizes the same statistical and system resource metrics as FedCSGA and selects the top-$m$ clients by ranking. This baseline isolates the effect of greedy ranking versus GA-based selection.
\end{itemize}

\subsubsection{Datasets and model architecture}
We evaluated the performance of FedCSGA on four benchmark datasets that are widely adopted in FL research.
\begin{itemize}
    \item \textbf{\textit{FMNIST}} \cite{xiao2017fashion} A dataset of grayscale images with dimensions \(28 \times 28 \times 1\), consisting of 10 classes, with 60,000 training samples and 10,000 test samples.
    \item \textbf{\textit{CIFAR-10}} \cite{krizhevsky2009learningcifar10} A dataset of color images with dimensions \(32 \times 32 \times 3\), containing 10 classes, with 50,000 training samples and 10,000 test samples.
    \item \textbf{\textit{CIFAR-100}} \cite{krizhevsky2009learningcifar10} A dataset of color images with dimensions \(32 \times 32 \times 3\), featuring 100 classes, with 50,000 training samples and 10,000 test samples.
    \item \textbf{\textit{Tiny-Imagenet}} \cite{le2015tinyimagenet} A dataset of color images with dimensions \(64 \times 64 \times 3\), comprising 200 classes, with 100,000 training samples, 10,000 validation samples, and 10,000 test samples.
\end{itemize}

For FMNIST, CIFAR-10, and CIFAR-100 datasets, we use the same CNN architecture used in \cite{mai2024rflpa,shao2024selective}. For the more complex Tiny-ImageNet dataset, a ResNet-18 model \cite{he2016deep} was used. For the FMNIST and Tiny-ImageNet datasets, each experiment was run for \(300\) communication rounds, whereas for CIFAR-10 and CIFAR-100, \(200\) rounds were used. Unless stated otherwise, a subset of \(10\) clients from a total of \(100\) was selected to participate in each round. All experiments were repeated three times with different random seeds. Following the protocol of \cite{masud2025stabilizing}, the final performance is computed as the average over the last 10 communication rounds, and we report the mean and standard deviation across runs. To emulate the client-server architecture and heterogeneous client devices, we implemented the FedCSGA algorithm along with all other baselines using the Flower framework~\cite{KUNDROO2025102276} and PyTorch~\cite{ansel2024pytorch}. We performed all of our experiments on an Intel Core i9-10900 3.70-GHz processor with 192 GB of RAM, and \(4 \times\) NVIDIA GeForce RTX 4090 GPUs.

\subsubsection{Heterogeneity settings}
To simulate realistic FL environments while maintaining conceptual simplicity, we model statistical and system heterogeneity with equal weighting by setting \(\lambda_{r} = 0.5\) and \(\lambda_{s} = 0.5\). 
For statistical heterogeneity, the training datasets were partitioned among 100 clients using a Dirichlet distribution parameterized by \( \alpha \), where smaller values of \( \alpha \) correspond to higher degrees of statistical heterogeneity (data non-IIDness) \cite{Selo_Kundroo_Kim_2025}. We consider three configurations: IID (uniform distribution), non-IID with \( \alpha = 0.3 \), and non-IID with \( \alpha = 0.6 \).
To emulate system heterogeneity in resource-constrained environments, we adopt settings similar to those used in \cite{fedprox_li,singhal2024greedy,zhao2022protea}, where clients are randomly assigned either (i) $1$ local epoch, a batch size of $8$, and a learning rate \( \eta = 0.001 \), or (ii) $3$ local epochs, a batch size of $32$, and a learning rate \( \eta = 0.01 \). In each communication round, clients were randomly assigned to one of these settings to simulate varying system capabilities. For the more complex Tiny-ImageNet dataset, the learning rates are adjusted to \( \eta = 0.00001 \) and \( \eta = 0.0001 \) for (i) and (ii), respectively.

\subsubsection{Genetic algorithm parameters}
To perform the client selection process, we tuned the hyperparameters of GA using grid search, the results of which are discussed in Section~\ref{subsec:ga_hyper_result}. During tuning, we monitored not only test accuracy but also the server-side runtime overhead for training each generation. Our hyperparameter selection accounted for both accuracy maximization and runtime minimization. A detailed analysis supporting these observations, including the communication and computation trade-offs across different configurations, is presented in Section~\ref{subsec:computationalanalysis}.
Through experimentation, we identified the best performing configuration with the total number of generations \(G=30\), population size \(P=30\), parent chromosome fraction \(\rho_p=0.5\), and child chromosome fraction \(\rho_c=0.5\). We incorporated the concept of elitism \cite{ahn2003elitism} and tournament selection \cite{miller1995genetic} for parent and child chromosomes, which ensures the preservation of an optimal subset while also maintaining diversity in the population. For the history-aware selection mechanism in FedCSGA, we found that moderate values of \( \gamma \) yield stable performance across datasets, while very large values excessively penalize frequently selected clients, therefore, \( \gamma = 0.03 \) was used in all reported results.

\begin{table}[!b]
\centering
\caption{Comparison of FedCSGA and baseline accuracies on four datasets under non-IID settings ($\alpha = \{0.3, 0.6\}$).}
\label{tab:main_results_noniid}
\resizebox{\textwidth}{!}{
\begin{tabular}{l|cc|cc|cc|cc}
\toprule
\multirow{2}{*}{\textbf{Algorithm}} & \multicolumn{2}{c|}{\textbf{FMNIST}} & \multicolumn{2}{c|}{\textbf{CIFAR-10}} & \multicolumn{2}{c|}{\textbf{CIFAR-100}} & \multicolumn{2}{c}{\textbf{Tiny-Imagenet}} \\
\cmidrule(lr){2-3} \cmidrule(lr){4-5} \cmidrule(lr){6-7} \cmidrule(lr){8-9}
 & $\boldsymbol{\alpha} = $ \textbf{0.3} & $\boldsymbol{\alpha} = $ \textbf{0.6} & $\boldsymbol{\alpha} = $ \textbf{0.3} & $\boldsymbol{\alpha} = $ \textbf{0.6} & $\boldsymbol{\alpha} = $ \textbf{0.3} & $\boldsymbol{\alpha} = $ \textbf{0.6} & $\boldsymbol{\alpha} = $ \textbf{0.3} & $\boldsymbol{\alpha} = $ \textbf{0.6} \\ 
\midrule
\textbf{FedAvg}     & 79.5 ± 2.3 & 82.0 ± 1.8 & 49.3 ± 3.1 & 55.4 ± 1.9 & 26.2 ± 0.7 & 27.2 ± 0.8 & 22.6 ± 0.8 & 22.8 ± 0.9 \\ 
\textbf{FedProx}    & 79.6 ± 1.1 & 81.8 ± 1.4 & 50.4 ± 2.8 & 56.1 ± 1.8 & 26.3 ± 0.6 & 27.2 ± 1.0 & 22.5 ± 0.6 & 23.3 ± 0.5 \\ 
\textbf{PowD}       & 80.9 ± 1.7 & 82.5 ± 1.4 & 51.3 ± 4.3 & 57.2 ± 1.4 & 25.6 ± 0.8 & 26.8 ± 0.7 & 22.7 ± 0.5 & 23.2 ± 0.9 \\
\textbf{AFL}        & 78.6 ± 2.6 & 82.0 ± 1.5 & 51.6 ± 2.9 & 56.8 ± 1.8 & 26.4 ± 0.7 & 27.2 ± 0.5 & 22.6 ± 0.4 & 22.7 ± 0.3 \\
\textbf{GreedyFed}  & 80.1 ± 1.3 & 83.3 ± 0.5 & 49.2 ± 1.3 & 52.3 ± 1.1 & 18.0 ± 0.2 & 18.1 ± 0.7 & 14.1 ± 0.2 & 14.1 ± 0.9 \\
\textbf{GreedyFL}  & \textbf{82.2 ± 2.2} & \textbf{84.9 ± 1.2} & 54.1 ± 2.3 & 58.4 ± 1.5 & 26.9 ± 0.4 & 27.2 ± 0.3 & 23.7 ± 0.6 & 24.3 ± 0.6 \\
\midrule
\textbf{FedCSGA}    & 81.9 ± 1.9 & 84.4 ± 1.3 & \textbf{55.6 ± 2.8} & \textbf{60.3 ± 2.1} & \textbf{29.9 ± 0.6} & \textbf{30.5 ± 0.4} & \textbf{29.1 ± 0.4} & \textbf{29.8 ± 0.2} \\
\bottomrule
\end{tabular}
}
\end{table}

\begin{figure}[!t]
    \centering
    \includegraphics[width=\textwidth]{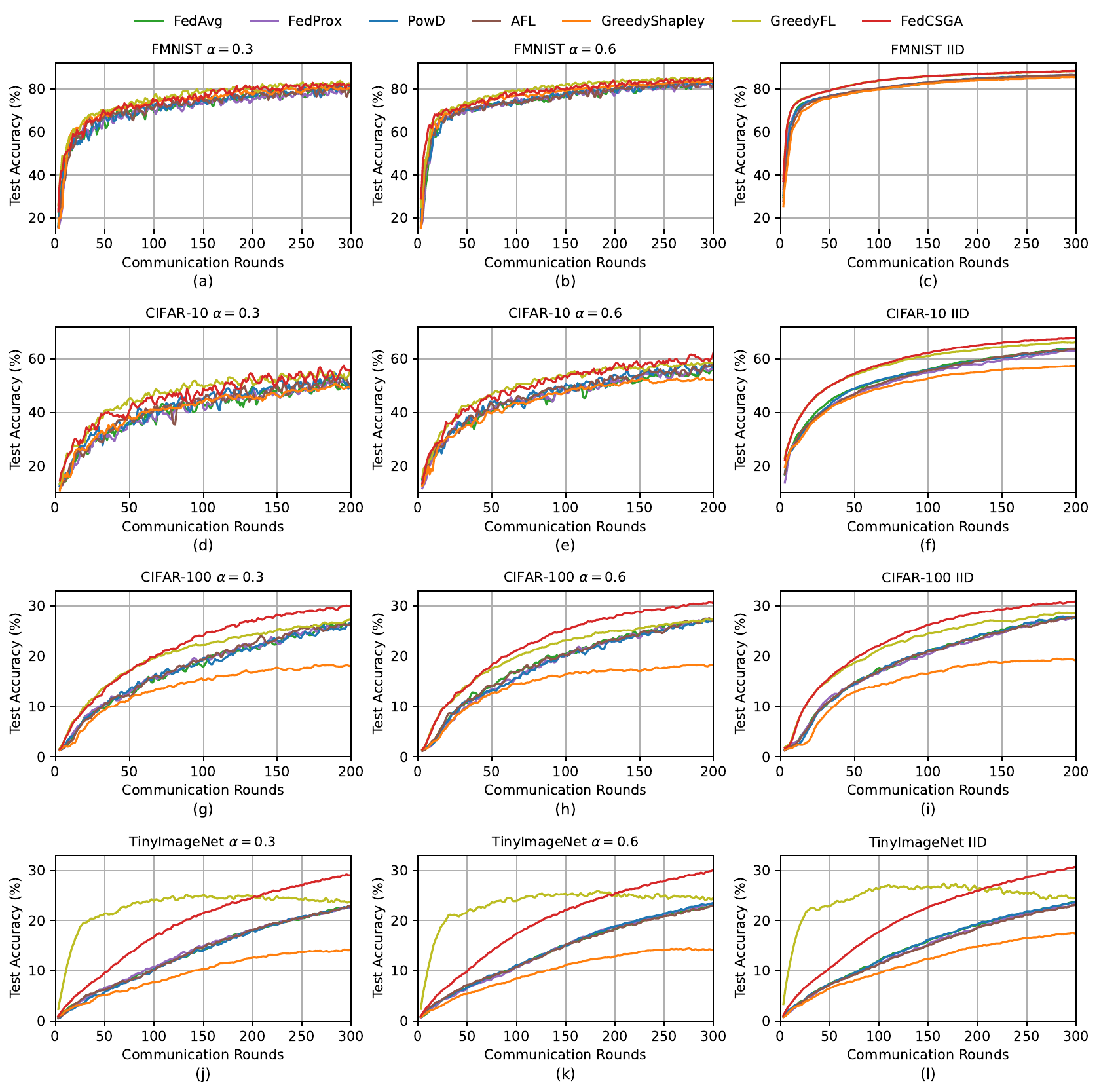}
    \caption{Performance comparison across FMNIST, CIFAR-10, CIFAR-100, and Tiny-Imagenet under IID and non-IID settings ($\alpha = \{0.3, 0.6\}$).}
    \label{fig:all_datasets}
\end{figure}

\begin{table}[!t]
\centering
\caption{Comparison of FedCSGA and baseline accuracies on four datasets under IID settings.}
\label{tab:main_results_iid}
\resizebox{0.65\textwidth}{!}{
\begin{tabular}{l|c|c|c|c}
\toprule
\textbf{Algorithm} & \textbf{FMNIST} & \textbf{CIFAR-10} & \textbf{CIFAR-100} & \textbf{Tiny-Imagenet} \\
\midrule
\textbf{FedAvg}     & 86.2 ± 0.1 & 63.7 ± 0.5 & 27.7 ± 0.5 & 23.6 ± 0.6 \\
\textbf{FedProx}    & 86.1 ± 0.3 & 63.0 ± 0.4 & 27.5 ± 0.3 & 23.0 ± 0.9 \\
\textbf{PowD}       & 86.5 ± 0.1 & 63.6 ± 0.3 & 27.7 ± 0.5 & 23.5 ± 0.5 \\
\textbf{AFL}        & 86.4 ± 0.2 & 63.5 ± 0.2 & 27.6 ± 0.3 & 22.9 ± 0.4 \\
\textbf{GreedyFed}  & 85.5 ± 0.3 & 57.3 ± 0.7 & 19.3 ± 0.2 & 17.4 ± 0.9 \\
\textbf{GreedyFL}  & 88.2 ± 0.1 & 66.2 ± 0.8 & 28.5 ± 0.2 & 24.5 ± 0.53 \\
\midrule
\textbf{FedCSGA}    & \textbf{88.3 ± 0.2} & \textbf{67.7 ± 0.6} & \textbf{30.7 ± 0.2} & \textbf{30.6 ± 0.2} \\
\bottomrule
\end{tabular}
}
\end{table}

\subsection{Performance comparison}\label{subsec:sota_result}
Table~\ref{tab:main_results_noniid} reports the mean and standard deviation of test accuracy for all the datasets across non-IID settings. For FMNIST, GreedyFL attains the highest mean accuracy at both \( \alpha = 0.3 \) and \( \alpha = 0.6 \), marginally outperforming FedCSGA (Figure~\ref{fig:all_datasets}, subplots (a) and (b)). Nevertheless, the performance gap remains within \(0.5\%\), and as dataset complexity increases, FedCSGA consistently yields larger and more stable improvements across baselines, indicating stronger scalability under more challenging non-IID conditions.
On CIFAR-10, FedCSGA exceeds the performance of AFL by \(4.0\%\) in the highly non-IID scenario (\( \alpha = 0.3 \)) and outperforms PowD by \(3.1\%\) under \( \alpha = 0.6 \) (Figure~\ref{fig:all_datasets}, subplots (d) and (e)). For the more complex CIFAR-100 dataset, where baseline performances are generally lower, FedCSGA achieves consistent improvements, surpassing AFL by \(3.5\%\) under \( \alpha = 0.3 \) and AFL by \(3.3\%\) under \( \alpha = 0.6 \) (Figure~\ref{fig:all_datasets}, subplots (g) and (h)). In case of Tiny-Imagenet, we see a significant improvement of \(6.4\%\) under \( \alpha = 0.3 \) and \(6.5\%\) under \( \alpha = 0.6 \) against PowD and FedProx, respectively (Figure~\ref{fig:all_datasets}, subplots (j) and (k)). Unlike GreedyFed, which suffers substantial performance drops on the more challenging CIFAR-100 and Tiny-Imagenet datasets, FedCSGA demonstrates faster convergence and maintains steady accuracy improvements with reduced variance across all settings. The performance degradation of GreedyFed may stem from its reliance on server-side validation data for Shapley Value estimation, which becomes unreliable when clients hold unique or non-overlapping data classes in complex datasets and highly non-IID settings.

As shown by Table~\ref{tab:main_results_iid}, FedCSGA also exhibits consistent performance gains in the IID settings. For instance, it outperforms FedAvg by a significant margin of \(7.0\%\) on Tiny-Imagenet (Figure~\ref{fig:all_datasets}, subplot (l)). Similarly, we see an improvement of \(4.0\%\) and \(3.0\%\) against FedAvg on CIFAR-10 and CIFAR-100 (Figure~\ref{fig:all_datasets}, subplots (f) and (i)), respectively, and \(1.8\%\) against PowD on FMNIST (Figure~\ref{fig:all_datasets}, subplot (c)). Compared to other baselines, these results highlight the joint consideration of statistical and system heterogeneity by FedCSGA, which ensures performance improvement even in the IID scenarios where the data across all clients follows a similar data distribution. This balanced approach to handling diverse data distributions and resource considerations is critical in federated learning, ensuring effective generalization across datasets with varying complexity and heterogeneity. Overall, these results underscore the practical applicability of FedCSGA in diverse federated learning environments.

\begin{table}[!b]
\centering
\caption{Accuracy comparison of FedCSGA against baselines across four client participation ratios.}
\label{tab:cpr_comparison}
\resizebox{0.60\textwidth}{!}{
\begin{tabular}{l|c|c|c|c}
\toprule
\multirow{2}{*}{\textbf{Algorithm}} & \multicolumn{4}{c}{\textbf{Client Participation Ratio (CPR)}} \\ 
\cmidrule(lr){2-5}
 & \textbf{0.10} & \textbf{0.15} & \textbf{0.20} & \textbf{0.30} \\ 
\midrule
\textbf{FedAvg}     & 49.3 ± 3.1 & 52.7 ± 2.7 & 54.3 ± 1.7 & 56.0 ± 2.0 \\ 
\textbf{FedProx}    & 50.4 ± 2.8 & 54.1 ± 1.7 & 54.2 ± 1.8 & 55.6 ± 1.3 \\ 
\textbf{PowD}       & 51.3 ± 4.3 & 52.8 ± 2.7 & 55.0 ± 2.2 & 56.2 ± 1.7 \\ 
\textbf{AFL}        & 51.6 ± 2.9 & 54.5 ± 2.2 & 54.9 ± 2.0 & 55.8 ± 1.8 \\
\textbf{GreedyFed}  & 49.2 ± 1.3 & 51.5 ± 0.6 & 54.0 ± 1.0 & 53.8 ± 2.0 \\ 
\textbf{GreedyFL}  & 54.1 ± 2.3 & 56.5 ± 1.8 & 58.2 ± 1.4 & 60.3 ± 1.1 \\ 
\midrule
\textbf{FedCSGA}    & \textbf{55.6 ± 2.8} & \textbf{57.0 ± 2.7} & \textbf{58.6 ± 1.8} & \textbf{60.7 ± 1.7} \\ 
\bottomrule
\end{tabular}
}
\end{table}

\begin{figure}[!b]
    \centering
    \includegraphics[width=\textwidth]{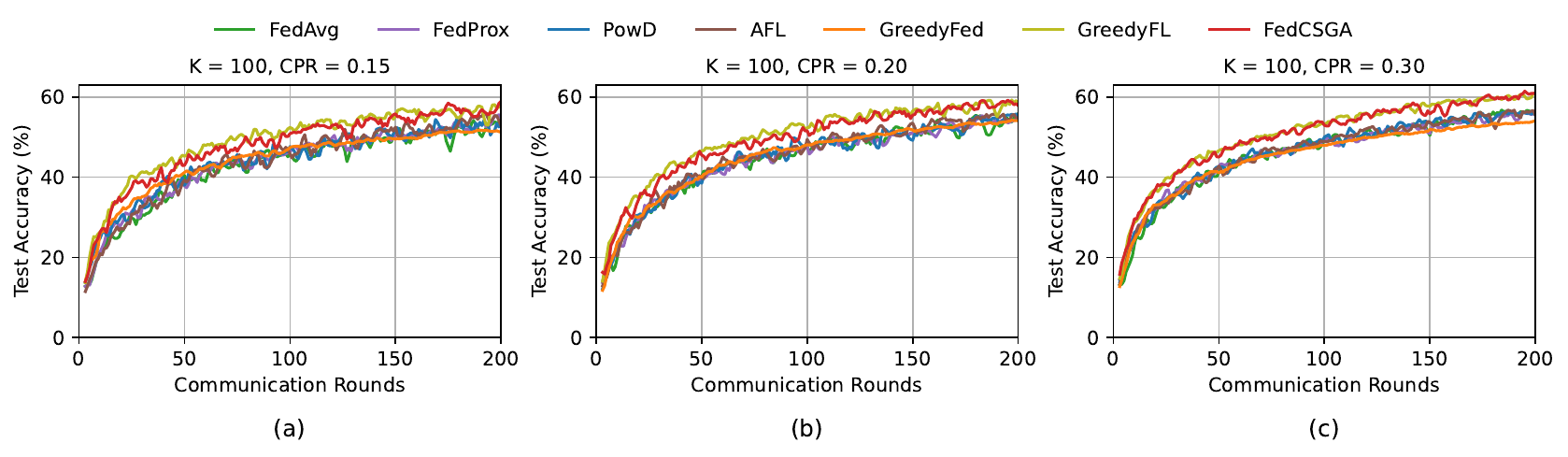}
    \caption{Performance comparison under varying client participation ratios (CPR = $0.15$, $0.20$, $0.30$).}
    \label{fig:cpr_graph}
\end{figure}

\subsection{Impact of partial client participation}\label{subsec:cpr_result}
To evaluate the robustness and adaptability of our proposed FedCSGA algorithm, we analyze its performance under varying client participation ratios (CPR) using the CIFAR-10 dataset with \( \alpha = 0.3 \). Table~\ref{tab:cpr_comparison} and Figure~\ref{fig:cpr_graph} illustrate the impact of different CPR values on FedCSGA and the baseline algorithms. As the number of participating clients per training round increases, all methods demonstrate improved performance, reflecting the general benefit of increased data diversity during training. Notably, FedCSGA consistently outperforms all baseline algorithms across varying CPR levels, underscoring its scalability and adaptability in complex FL environments. For instance, at CPR \(0.10\), FedCSGA achieves a significant improvement of \(1.5\%\) over the next best-performing baseline, GreedyFL. Overall, these findings reaffirm the effectiveness of FedCSGA in addressing the inherent challenges of partial client participation in FL, making it a practical choice for diverse and challenging real-world deployments.

\subsection{Sensitivity Analysis of GA Hyper-parameters}\label{subsec:ga_hyper_result}
This subsection investigates the sensitivity of FedCSGA to key GA hyper-parameters. Specifically, we analyze how the population size \( P \) and the number of generations \( G \) influence convergence behavior and model accuracy, as well as how the fractions of parent (\( \rho_p \)) and child (\( \rho_c \)) chromosomes affect the balance between solution exploitation and exploration. The results provide insights into the trade-offs among diversity, convergence speed, and performance, guiding the selection of effective GA configurations.

\subsubsection{Impact of hyper-parameters \texorpdfstring{$P$ and $G$}{P and G}}
\begin{figure}[!h]
    \centering
    \includegraphics[width=0.6\textwidth]{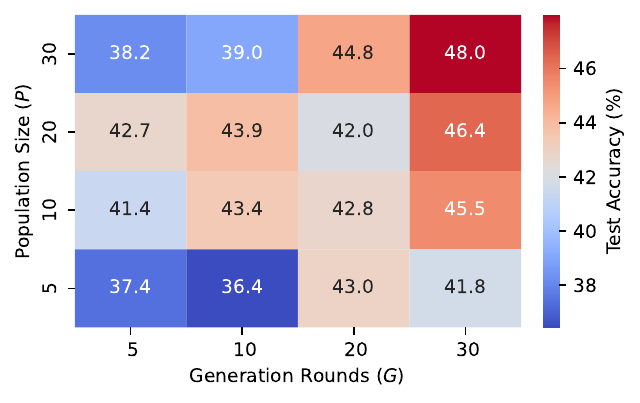}
    \caption{Impact of population size \(P\) and generations rounds \(G\).}
    \label{fig:population_generation}
\end{figure}

Figure~\ref{fig:population_generation} presents the performance of FedCSGA under varying values of population size \( P \) and generation rounds \( G \). The accuracy values are reported across \( G \in \{5, 10, 20, 30\} \) and \( P \in \{5, 10, 20, 30\} \). It can be observed that FedCSGA demonstrates suboptimal performance for smaller values of \( G \), as insufficient evolutionary rounds limit the algorithm's ability to evolve chromosomes toward high-fitness solutions effectively. Similarly, for smaller values of \( P \), the reduced diversity within the population restricts the GA's ability to explore a broad solution space, resulting in lower accuracy. For example, when \( P = 5 \) and \( G = 5 \), the accuracy is the lowest at \( 37.4\% \), highlighting the compounded limitations of a small population and minimal evolutionary rounds. Conversely, higher values of \( P \) and \( G \) result in noticeable performance gains. For instance, when \( P = 30 \) and \( G = 30 \), the accuracy peaks at \( 48.0\% \), indicating the advantages of increased diversity and extended evolution. Notably, FedCSGA also achieves fair performance within moderate ranges of these hyper-parameters, such as \( P = 20 \) and \( G = 10 \), with an accuracy of \( 43.9\% \). These findings emphasize the importance of balancing \( P \) and \( G \) to achieve effective performance. Since the performance improvement exhibits a clear correlation with increased computational resources, we recommend utilizing values around \( P \ge 20 \) and \( G \ge 10 \) as robust default configurations. This guideline allows practitioners to achieve reliable performance without extensive hyper-parameter tuning, thereby effectively mitigating sensitivity concerns in practical deployments.

\subsubsection{Impact of hyper-parameters \texorpdfstring{$\rho_p$}{rho\_p} and \texorpdfstring{$\rho_c$}{rho\_c}}
\begin{figure}[!t]
    \centering
    \includegraphics[width=0.6\textwidth]{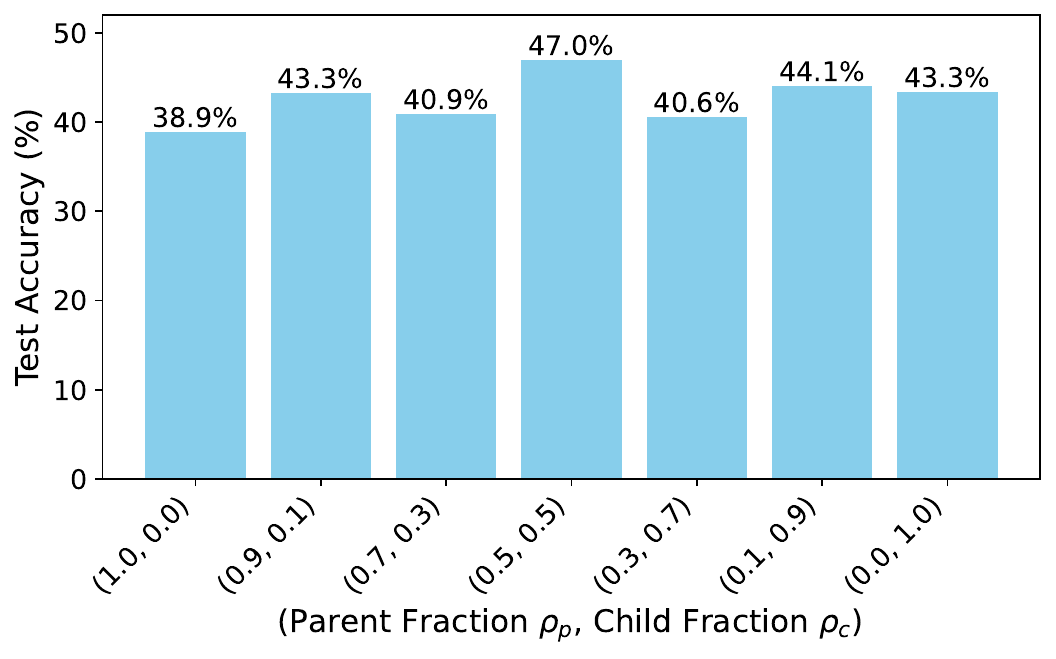}
    \caption{Impact of parent and child chromosome fractions (\(\rho_p, \rho_c\)).}
    \label{fig:parent_child_frac}
\end{figure}

Figure~\ref{fig:parent_child_frac} reports the test accuracy of FedCSGA across varying fractions of parent chromosomes (\( \rho_p \)) and children chromosomes (\( \rho_c \)). These hyper-parameters define the trade-off between retaining prior solutions and generating new ones. We enforce \( \rho_p + \rho_c = 1 \) to ensure that the total number of chromosomes in the population remains constant across generations. The result shows that excessively high values of both \( \rho_p \) and \( \rho_c \) leads to premature convergence. For high \( \rho_p \) values, this might occur due to the reduced diversity in the population, limiting the algorithm's ability to explore alternative client subsets in the search space. Conversely, very high values of \( \rho_c \) result in the loss of valuable high-performing solutions from earlier generations, yielding suboptimal accuracy. Robust performance is achieved when \( \rho_p \) and \( \rho_c \) are equal, allowing for a balanced approach between retaining strong parent chromosomes and introducing diverse child chromosomes. For instance, when \( \rho_p = 0.5 \) and \( \rho_c = 0.5 \), the test accuracy peaks at \( 47.0\% \), demonstrating the effectiveness of this balanced approach. These findings highlight the importance of tuning \( \rho_p \) and \( \rho_c \) for efficient client selection. By carefully balancing parent retention and children generation, FedCSGA can effectively leverage high-performing clients while maintaining population diversity to explore promising client subsets. Given that the performance sensitivity follows a predictable trend favoring balance, we recommend setting \( \rho_p \approx \rho_c \approx 0.5\) as a standard default configuration. Adopting this balanced setting ensures stable performance and prevents degradation caused by extreme parameter choices.

\subsection{Robustness to Privacy Constraints and System Unreliability}\label{subsec:robustness_analysis}

In practical edge-cloud FL environments, the assumptions of perfectly accurate metric reporting and fully synchronous client availability rarely hold. To address these limitations and evaluate the practical applicability of FedCSGA, we conduct a sensitivity analysis focusing on two critical real-world challenges: privacy-preserving noise injection and delayed metric reporting due to network unreliability.

\subsubsection{Threat Model and Privacy-Preserving Mitigation}
Although FedCSGA does not require the sharing of raw data or model parameters during the selection phase, the transmission of resource metrics ($\text{Res}_{\text{stat}}$ and $\text{Res}_{\text{sys}}$) can still leak sensitive metadata. Under an honest-but-curious server threat model, an adversary could infer a client's level of data collection engagement from its dataset size~\cite{melis2019exploiting}, or utilize the exact training loss to conduct membership inference attacks, as loss values often correlate with the presence of unique or anomalous data samples~\cite{nasr2019comprehensive}.

To mitigate this, we implement a Local Differential Privacy (LDP) mechanism where clients perturb their metrics prior to transmission. Specifically, clients inject zero-mean Laplace noise into their statistical metrics ($\text{Res}_{\text{stat}}$), parameterized by a noise scale $\beta$. Simultaneously, to simulate system unreliability and estimation errors (e.g., CPU thermal throttling or background tasks), we apply a multiplicative uniform noise $p$ to the system resource metric ($\text{Res}_{\text{sys}}$). As shown in Table~\ref{tab:privacy_system_noise}, we vary the privacy noise scale ($\beta \in[0.0, 3.0]$) and the system variance ($p \in [0.0, \pm0.8]$). For moderate perturbations (up to $\beta = 1.0$ and $p = \pm0.6$), the decrease in global model performance is negligible, demonstrating the inherent resilience of the proposed algorithm. This robustness is fundamentally tied to the population-based nature of the GA. Because the evolutionary process evaluates the relative fitness of an entire client subset rather than demanding absolute precision from individual clients, minor perturbations do not significantly alter the comparative ranking of the candidate chromosomes. Consequently, the algorithm consistently identifies better client combinations despite moderate metric obfuscation. Conversely, as $\beta$ and $p$ scale to extreme values ($\beta \in [2.0, 3.0], p \in [\pm0.7, \pm0.8]$), we observe a more pronounced degradation in performance. This aligns with well-documented trade-offs in differentially private FL, where excessive noise magnitudes overwhelm the true utility signals, causing the optimization landscape to flatten and forcing the selection process to degenerate toward random sampling \cite{kairouz2021advances}.

\begin{table}[!h]
\centering
\caption{Impact of privacy noise ($\beta$) and system variance ($p$) on FedCSGA performance.}
\label{tab:privacy_system_noise}
\resizebox{0.6\textwidth}{!}{
\begin{tabular}{c|c|c}
\toprule
\textbf{Privacy Noise ($\beta$)} & \textbf{System Variance ($p$)} & \textbf{Accuracy (\%)} \\
\midrule
0.0 & 0.0 & 55.6 ± 2.8 \\
0.3 & $\pm$ 0.4 & 54.3 ± 0.2 \\
0.5 & $\pm$ 0.5 & 53.3 ± 0.4 \\
1.0 & $\pm$ 0.6 & 51.3 ± 0.7 \\
2.0 & $\pm$ 0.7 & 48.8 ± 1.8 \\
3.0 & $\pm$ 0.8 & 47.0 ± 1.8 \\
\bottomrule
\end{tabular}
}
\end{table}

\subsubsection{Robustness to Asynchronous and Delayed Metrics}
Furthermore, edge devices frequently experience intermittent connectivity, resulting in asynchronous availability. To simulate this, we introduce a drop probability ($P_{\text{stale}}$), representing the likelihood that a client fails to transmit its current resource metrics in a given round. In such events, the server relies on the cached, delayed metrics from the client's last successful transmission. 

Table~\ref{tab:staleness_effect} illustrates the impact of increasing $P_{\text{stale}}$ on global model accuracy. Remarkably, FedCSGA maintains near-baseline performance even when a significant majority of the client pool fails to report current metrics. At an extreme drop probability of $P_{\text{stale}} = 0.8$, where $80\%$ of selection decisions rely on delayed metrics, the test accuracy experiences a negligible decline to $55.1\%$. This robustness can be attributed to the temporal locality of FL dynamics, where client dataset sizes remain static, and local training loss decreases smoothly over consecutive rounds. Consequently, delayed metrics serve as a highly reliable proxy for the client's current utility, ensuring that transient network failures do not derail the FedCSGA optimization process.

\begin{table}[!ht]
\centering
\caption{Impact of delayed client metrics (staleness probability $P_{\text{stale}}$) on FedCSGA performance.}
\label{tab:staleness_effect}
\resizebox{0.5\textwidth}{!}{
\begin{tabular}{c|c}
\toprule
\textbf{Staleness Probability ($P_{\text{stale}}$)} & \textbf{Accuracy (\%)} \\
\midrule
0.20 & 55.9 ± 0.5 \\
0.40 & 55.8 ± 0.4 \\
0.60 & 55.7 ± 1.3 \\
0.80 & 55.1 ± 0.8 \\
\bottomrule
\end{tabular}
}
\end{table}

\subsection{Ablation study on fitness function}
We conduct an ablation study to analyze the contribution of resource-aware metrics and the proposed history-aware weighting factor in the FedCSGA fitness formulation. Table~\ref{tab:ablation_fedcsga} reports the test accuracy on CIFAR-10 ($\alpha = 0.6$) under three configurations: (i) using only the resource-aware metric $\text{Res}(c)$, which aggregates normalized statistical and system indicators, (ii) using only the history-aware weighting factor $W(c)$, where clients are selected by sorting them in ascending order of their past selection counts (i.e., prioritizing the least-selected clients), and (iii) the full FedCSGA configuration that jointly incorporates both components.
When only $\text{Res}(c)$ is used, the model achieves a test accuracy of $59.0\%$, indicating that prioritizing clients based on instantaneous statistical and system utility is effective for improving convergence and final performance. In contrast, relying solely on the history-aware factor yields significantly lower accuracy ($55.2\%$), as this configuration promotes participation diversity without explicitly considering data quality or system efficiency. The best performance is obtained when both $\text{Res}(c)$ and $W(c)$ are combined, achieving $60.3\%$ accuracy. This improvement demonstrates that historical participation information complements resource-aware selection by discouraging repeated selection of similar client subsets across rounds while preserving utility-driven selection.
These results validate the design of the proposed fitness function, showing that neither resource awareness nor historical weighting alone is sufficient to achieve efficient performance. Instead, their joint use enables FedCSGA to balance short-term training utility with long-term participation dynamics, resulting in more stable and effective client selection under heterogeneous federated learning settings.

\begin{table}[h]
    \centering
    \caption{Ablation study of FedCSGA showing the impact of resource-aware client selection $\text{Res}(c)$ and history-aware weighting $W(c)$.}
    \label{tab:ablation_fedcsga}
    \resizebox{0.7\textwidth}{!}{
    \begin{tabular}{c|c|c}
        \toprule
        \textbf{Resource Metrics} $\text{Res}(c)$ &
        \textbf{History-Aware Factor} $W(c)$ &
        \textbf{Accuracy (\%)} \\
        \midrule
        \checkmark &  & 59.0 ± 0.6 \\
         & \checkmark & 55.2 ± 0.7 \\
        \checkmark & \checkmark & \textbf{60.3 ± 2.1} \\
        \bottomrule
    \end{tabular}
    }
\end{table}

\subsection{Computational and communication complexity}\label{subsec:complexityanalysis}
This subsection analyzes the computational and communication costs associated with FedCSGA. We first derive the time complexity of the GA-based client selection process, highlighting how the population size, number of generations, and selected clients affect server-side computation. We then empirically evaluate the communication efficiency and per-round computation time of FedCSGA in comparison with baseline methods, illustrating the trade-off between additional server computation and reduced communication rounds.

\subsubsection{Time complexity analysis}\label{subsec:timealanalysis}
The time complexity of FedCSGA is primarily governed by the GA-based client selection mechanism. In each communication round, the algorithm first initializes a population of $P$ random client subsets (chromosomes), each containing $m$ clients. Random sampling of $m$ clients from a total of $K$ can be implemented in expected $O(m)$ time using efficient sampling procedures. Thus, initializing the population of $|P|$ chromosomes and evaluating their fitness together incur an overall cost of $O(|P| m)$. Fitness evaluation of each chromosome is performed by summing the resource metrics of its $m$ clients, which takes $O(m)$ time per chromosome. During each of the $G$ generations, the GA performs parent selection, crossover, and mutation. Parent selection identifies the top $\rho_p |P|$ chromosomes based on the fitness value, which can be performed in $O(|P|log|P|)$ time using sorting, or in expected $O(|P|)$ time using efficient selection algorithms such as QuickSelect~\cite{hoare1961find}. Crossover operations produce approximately $\rho_c |P|$ child chromosomes of size $m$ by recombining segments from two parent chromosomes. Each crossover involves copying $O(m)$ elements, resulting in $O(\rho_c |P|  m)$ time per generation. Mutation modifies each of the $\rho_c |P|$ children by replacing clients with a small probability ($1/m$), again requiring $O(\rho_c |P|  m)$ time. The cost of updating the population, which includes merging parents and children chromosomes, is negligible at $O(|P|)$. As such, the per-generation cost is dominated by crossover and mutation, both scaling as $O(|P|  m)$. Aggregating over $G$ generations, the total complexity of the GA process becomes $O( G |P| m)$. The one-time initialization cost does not affect the overall asymptotic complexity. Thus, the total time complexity of the FedCSGA client selection algorithm is $O(G |P| m)$. Given that the selected subset size $m$ is typically much smaller than the total number of clients $K$ ($m \ll K$), the complexity is effectively linear in $m$. If $m$ were on the order of $K$, the complexity would scale as $O(G |P| K)$.

\subsubsection{Communication complexity analysis}\label{subsec:computationalanalysis}
We also evaluate the communication and computation complexity of FedCSGA on CIFAR-10 ($\alpha = 0.3$) with other baselines using two metrics: \(R_\text{50}\), the number of communication rounds required to achieve a test accuracy of \(50\%\), and \(T_\text{comp}\), the average computation time (in seconds) per round. The computation time includes the server's client selection time and the time taken by selected clients to perform local updates.
As shown in Table~\ref{tab:comm_comp_complexity}, FedCSGA achieves a significant reduction in the number of communication rounds while maintaining a comparable computation time per round to simpler baselines such as FedAvg and FedProx. This demonstrates that, under moderate GA settings, the evolutionary search achieves efficient client selection without noticeable computational burden.

\begin{table}[!th]
    \centering
    \caption{Comparison of \( R_{50} \), and \( T_{\text{comp}} \) (sec) for different client participation ratios.}
    \label{tab:comm_comp_complexity}
    \resizebox{0.55\textwidth}{!}{
    \begin{tabular}{l|cc|cc|cc}
        \toprule
        \multirow{3}{*}{\textbf{Algorithm}} 
        & \multicolumn{6}{c}{\textbf{Client Participation Ratio (CPR)}} \\ 
        \cmidrule{2-7}
         & \multicolumn{2}{c|}{\textbf{0.10}} 
         & \multicolumn{2}{c|}{\textbf{0.20}} 
         & \multicolumn{2}{c}{\textbf{0.30}} \\ 
        \cmidrule{2-7}
         & \(\boldsymbol{R_\text{50}}\) & \(\boldsymbol{T_\text{comp}}\) & \(\boldsymbol{R_\text{50}}\) & \(\boldsymbol{T_\text{comp}}\) & \(\boldsymbol{R_\text{50}}\) & \(\boldsymbol{T_\text{comp}}\) \\ 
        \midrule
        \textbf{FedAvg}     & 113 & 54 & 95 & 76 & 87 & 79 \\ 
        \textbf{FedProx}    & 112 & 54 & 99 & 55 & 92 & 57 \\ 
        \textbf{PowD}       & 100 & 58 & 89 & 93 & 84 & 97 \\ 
        \textbf{AFL}        & 107 & 54 & 94 & 55 & 81 & 56 \\ 
        \textbf{GreedyFed}  & 165 & 282 & 115 & 735 & 118 & 1573 \\ 
        \textbf{GreedyFL}  & 68 & 54 & 59 & 55 & 63 & 56 \\ 
        \midrule
        \textbf{FedCSGA}    & 64 & 64 & 65 & 72 & 57 & 78 \\
        \bottomrule
    \end{tabular}
    }
\end{table}

However, when client participation ratio increases, although FedCSGA continues to reduce the communication rounds required to reach the target accuracy, the computation time per round increases slightly due to the increased number of fitness evaluations performed at the server during client selection. Despite this, the additional server-side computation leads to a substantially faster overall convergence, which effectively lowers the total communication cost, which is the dominant factor in FL efficiency. In many real-world FL deployments, particularly in cross-device and bandwidth-limited scenarios, network communication is orders of magnitude slower and more expensive than server computation. Therefore, investing a few additional seconds of server processing to make more informed client selections yields a net reduction in total training time and communication expenditure.

Overall, the comprehensive evaluation across multiple datasets and heterogeneous settings confirms the overall efficacy of FedCSGA. The empirical findings highlight its superior performance, accelerated convergence speed, and notable robustness against both privacy-preserving noise and system-level delays. While these results validate the algorithmic strengths of the proposed approach under a wide array of simulated conditions, transitioning this framework into practical, real-world deployments necessitates the careful consideration of several inherent design trade-offs, which we explore in the subsequent section.

\section{Discussion and Limitations}
\label{sec:discussion}

This section further discusses the practical design trade-offs associated with deploying FedCSGA in real-world federated learning environments. Specifically, we examine the balance between convergence efficiency and long-term client fairness, the systemic assumptions required for resource reporting, and the scalability limitations inherent to massive cross-device deployments.

Concerning the balance between convergence efficiency and selection bias, FedCSGA does not aim to enforce strict long-term fairness guarantees. Instead, it achieves a soft participation balance through its history-aware decay mechanism, which prevents the exploitation of frequently selected clients. While prioritizing high-loss clients accelerates early convergence, this approach effectively mitigates persistent selection bias without relying on rigid, per-client frequency heuristics.

Beyond algorithmic fairness, deploying FedCSGA introduces specific system-level dependencies. Foremost, the framework relies on the assumption that clients transmit small resource summaries to the server at each training round. While this introduces a communication overhead proportional to the active candidate pool, the transmitted metadata (e.g., loss values and processing times) is extremely lightweight and remains negligible compared to the bandwidth required for exchanging model parameters.

This reliance on per-round metric reporting naturally dictates the scalability boundaries of the framework. Currently, FedCSGA is optimized for cross-silo and medium-scale cross-device FL deployments. In these settings, the server-side GA computation remains highly manageable, as its operational cost is tied to the selected client subset size rather than the total client pool. However, scaling this framework to environments where $K$ reaches millions of concurrent clients would expose limitations in gathering and processing per-round metrics. In such extreme scales, FedCSGA would require supplementary architectural modifications, such as hierarchical pre-selection or decentralized metric caching, to maintain its efficiency. 

Overall, the above analysis highlights the key design trade-offs in FedCSGA, demonstrating how the framework balances convergence efficiency, robustness, and system-level considerations in practical deployments.

\section{Conclusion}\label{sec:conclusion}
This paper presented FedCSGA, a resource-aware client selection framework for federated learning that explicitly addresses statistical and system heterogeneity while accounting for long-term participation dynamics. FedCSGA formulates client selection as a subset-level optimization problem and employs a genetic algorithm to jointly consider resource metrics and historical participation effects, avoiding reduction to per-client greedy ranking. Extensive experimental results across heterogeneous settings demonstrate improved performance and empirically stable convergence under moderate computational overhead. 
The effectiveness of FedCSGA stems from its ability to adapt client participation over training rounds by discouraging repeated selection of similar client subsets, thereby balancing utility-driven selection with participation diversity. While the current design prioritizes robustness, interpretability, and feasibility, several extensions remain open. Future work will investigate communication-efficient metric reporting strategies for large-scale cross-device deployments, privacy-preserving resource aggregation mechanisms, and more specialized genetic operators to further improve efficiency and scalability under extreme heterogeneity.

\section*{Declaration of competing interest}
The authors declare that they have no known competing financial interests or personal relationships that could have appeared to influence the work reported in this paper.

\section*{Acknowledgments}
This work was supported by the Institute of Information \& communications Technology Planning \& Evaluation (IITP) grant funded by the Korean government (MSIT) (RS-2025-02217801, Development of core technologies for integrated management of large-scale on-device AI objects and networks).

\bibliographystyle{elsarticle-num} 
\bibliography{references}

\end{document}